# When PCOS Meets Eating Disorders: An Explainable AI Approach to Detecting the Hidden Triple Burden

Apoorv Prasad
*Computer Science, College of Engineering & Applied Science*
*University of Wisconsin - Milwaukee*
Milwaukee, USA
prasada@uwm.edu

Susan McRoy
*Computer Science, College of Engineering & Applied Science*
*University of Wisconsin - Milwaukee*
Milwaukee, USA
mcroy@uwm.edu

***Abstract*—Women with polycystic ovary syndrome (PCOS) face substantially elevated risks of body image distress, disordered eating, and metabolic challenges, yet existing natural language processing approaches for detecting these conditions lack transparency and cannot identify co-occurring presentations. We developed small, open-source language models to automatically detect this triple burden in social media posts with grounded explainability. We collected 1,000 PCOS-related posts from six subreddits, with two trained annotators labeling posts using guidelines operationalizing Lee et al. (2017) clinical framework. Three models (Gemma-2-2B, Qwen3-1.7B, DeepSeek-R1-Distill-Qwen-1.5B) were fine-tuned using Low-Rank Adaptation to generate structured explanations with textual evidence. The best model achieved 75.3 percent exact match accuracy on 150 held-out posts, with robust comorbidity detection and strong explainability. Performance declined with diagnostic complexity, indicating their best use is for screening rather than autonomous diagnosis.**

## I. Introduction

Polycystic ovary syndrome (PCOS) affects 6-12% of reproductive-age women and is associated with significant metabolic, reproductive, and psychological comorbidities. A large-scale systematic review and meta-analysis drawing on 20 studies and over 28,000 women with PCOS found elevated odds of any eating disorder diagnosis (OR 1.53; 95% CI 1.29, 1.82), with specifically increased odds of BED, BN, and disordered eating, but notably no increased risk of anorexia nervosa [1]. Critically, elevated disordered eating scores persisted regardless of BMI [1]. Single-center data further report prevalence rates of 17.6% for BED, 6.1% for BN, and 12.9% for night eating syndrome in women with PCOS, alongside rates of body image distress substantially exceeding controls (shape concerns: 41.22% vs. 13.33%; weight concerns: 31.76% vs. 9.52%) [2]. This "triple burden" of co-occurring body image distress, disordered eating behaviors, and metabolic disorders presents unique clinical management challenges: current guidelines recommend weight loss for metabolic management [3], yet these recommendations may exacerbate or conflict with eating disorder treatment protocols [4]. Meanwhile, insulin resistance and weight loss resistance further complicate eating disorder recovery [2], and individuals with BED experience less weight loss, more rapid weight regain, and higher treatment attrition [5].

Psychological distress in PCOS extends beyond eating pathology. Women with PCOS show substantially elevated rates of anxiety (41.22% vs. 16.98%) and depression (12.16% vs. 3.77%) even after controlling for BMI [2], with the presence of anxiety conferring an almost six-fold increased risk of abnormal eating disorder screening scores. Eating disorder symptoms correlate inversely with health-related quality of life across all domains, most strongly in emotional wellbeing ($r = -0.513$) and weight-related quality of life ($r = -0.799$) [2], suggesting that undetected eating pathology may undermine other interventions.

Despite these risks, screening remains inadequate. While Australian guidelines recommend screening all women with PCOS for anxiety, depression, and disordered eating [3], implementation is limited. Structured instruments like the Eating Disorder Examination-Questionnaire (EDE-Q) are resource-intensive and may miss individuals who do not recognize their symptoms or who experience stigma around disclosing eating behaviors.

Online peer support communities represent an increasingly recognized resource for understanding PCOS-related psychological distress [6]. Research demonstrates that online health communities facilitate greater self-disclosure than face-to-face clinical settings [8], and women freely discuss body image, eating behaviors, and the emotional toll of PCOS across platforms including Reddit, Instagram, and TikTok [7,9,10]. These discussions capture lived experiences of the triple burden in ways structured assessments may miss [11,12]. However, manual analysis at scale is infeasible, and recent advances in transformer-based language models have demonstrated remarkable capabilities in understanding clinical text [14] and identifying nuanced patterns in mental health discourse [15], creating new opportunities for systematic analysis of patient-generated health data [13].

Existing NLP work faces three critical gaps. First, most mental health NLP relies on large proprietary models that function as "black boxes" [16], introducing reproducibility [18], data governance [19], and clinical safety concerns [17]. A GPT-3.5-based system applied to PCOS achieved 75.8% diagnostic accuracy and 96.9% sensitivity on structured clinical records [33], yet relied on proprietary infrastructure and was not evaluated on the naturalistic, patient-generated text where psychological comorbidities are most freely expressed. Second, existing systems target isolated conditions such as depression, anxiety, or suicidality [20], failing to capture the interconnected nature of distress in chronic illness. In PCOS, body image distress, disordered eating, and metabolic challenges interact bidirectionally [2,21], yet no NLP system simultaneously detects and characterizes their co-occurrence [22]. Third, clinical explainability remains underexplored: while recent work has applied attention visualization, saliency mapping, and feature importance techniques to mental health NLP models [23], these post-hoc methods often fail to align with clinical reasoning [24], and without justifications grounded in established diagnostic frameworks, even accurate models remain unsuitable for clinical decision support [25].

This study addresses all three gaps. We fine-tune three small, open-source language models (DeepSeek-R1-Distill-Qwen-1.5B, Qwen3-1.7B, Gemma-2-2B) on 700 annotated PCOS-related Reddit posts to simultaneously detect body image distress, disordered eating, and metabolic challenges, while generating structured, evidence-grounded justifications aligned with the clinical framework of Lee et al. (2017) [2]. By demonstrating that compact, transparent models can achieve screening-suitable performance without proprietary infrastructure, this work provides a foundation for accessible, ethical, and meaningful AI support for ill populations.

## II. Materials & Methods

### A. Data Collection

We collected data from six subreddits spanning eating disorder-focused and weight management communities: r/AnorexiaNervosa, r/BingeEatingDisorder, r/bulimia, r/EatingDisorders, r/EDAnonymous, and r/loseit. Our selection strategy deliberately included both disorder-specific subreddits (focusing on restrictive eating, binge eating, and binge-purge behaviors) and broader eating disorder communities to capture the full spectrum of eating disorder presentations, including diagnostically ambiguous cases, subclinical presentations, and individuals experiencing symptoms across multiple categories. The inclusion of r/loseit, a weight management community, serves an important comparative function, enabling us to distinguish between normative weight management concerns and clinically significant eating disorder behaviors. This is particularly relevant given that eating behaviors exist on a continuum from functional to dysfunctional patterns, with research showing that up to 48 percent of individuals exhibit disordered eating when conceptualized along this spectrum [26], making r/loseit valuable for understanding the transition zone between normative weight loss efforts and eating behaviors.

We employed a two-stage sampling strategy combining keyword-based filtering with random sampling. Using the Reddit PRAW API, we first extracted all posts mentioning "PCOS" (case-insensitive) from the selected subreddits, yielding 1,800 posts. This keyword-based approach was necessary to focus our dataset on individuals with polycystic ovary syndrome, given established literature linking PCOS with increased risk of eating disorders, body image disturbances, and metabolic challenges [2]. From this pool, we randomly sampled 1,000 posts to create our final dataset. Random sampling ensures our dataset is not biased toward any particular subreddit or post characteristics, allows us to characterize prevalence of different content types within the broader population of PCOS-related posts, and creates a manageable dataset size balancing sufficient examples against practical constraints of manual annotation. We acknowledge that requiring explicit mention of "PCOS" excludes posts from individuals with the condition who do not explicitly identify themselves, which could occur if users discuss PCOS-related symptoms without using the term itself. However, this limitation is acceptable for our research objectives, as we aim to study how individuals who explicitly identify as having PCOS describe their experiences, rather than attempt to infer PCOS status from symptom descriptions alone, which would introduce substantial diagnostic uncertainty.

Following data collection, two trained annotators independently labeled all 1,000 posts according to the annotation framework described above. The annotated dataset was then split using stratified sampling based on the number of co-occurring conditions to ensure balanced representation across splits: training set (700 posts, 70%), validation set (150 posts, 15%), and test set (150 posts, 15%). Overall, 22.4% of posts (n=224) exhibited body image distress, 20.4% (n=204) showed disordered eating behaviors, and 37.6% (n=376) described metabolic or weight management challenges (Table II). The stratification procedure successfully maintained similar proportions across all splits, with distributions varying by less than 4 percentage points between splits for each construct. Regarding co-occurrence patterns as shown in Table I, 39.5% of posts (n=395) contained no target constructs, 43.4% (n=434) exhibited one construct, 14.5% (n=145) showed two co-occurring constructs, and 2.6% (n=26) demonstrated all three constructs simultaneously. This distribution was consistent across training (39.6%, 43.1%, 14.7%, 2.6%), validation (39.3%, 43.3%, 14.7%, 2.7%), and test sets (39.3%, 44.7%, 13.3%, 2.7%), confirming successful stratification.

TABLE I Distribution of co-occurring conditions

| Construct | Train (n=700) | Validation (n=150) | Test (n=150) | Overall (N=1,000) |
|---|---|---|---|---|
| **Co-occurring Conditions across sets** | | | | |
| *0 conditions* | 277 (39.6%) | 59 (39.3%) | 59 (39.3%) | 395 (39.5%) |
| *1 condition* | 302 (43.1%) | 65 (43.3%) | 67 (44.7%) | 434 (43.4%) |
| *2 conditions* | 103 (14.7%) | 22 (14.7%) | 20 (13.3%) | 145 (14.5%) |
| *3 conditions* | 18 (2.6%) | 4 (2.7%) | 4 (2.7%) | 26 (2.6%) |

All data collection and annotation procedures were conducted in accordance with institutional research ethics guidelines. As posts were collected exclusively from publicly accessible Reddit communities, no interaction with human subjects occurred and individual informed consent was not required. All usernames and account identifiers were removed prior to annotation, and annotators worked only with anonymized post text. The dataset is not publicly released in raw form to mitigate re-identification risks. This study was determined to be exempt from full IRB review as it involved only publicly available data with no collection of identifiable private information.

### B. Annotation Framework and Process

Our annotation scheme was grounded in the clinical framework established by Lee et al. [2], who demonstrated that women with PCOS had 4.75 times higher odds of abnormal eating disorder screening scores compared to controls, with clinically significant elevations in both shape concern and weight concern. Critically, this increased risk persisted after controlling for BMI, suggesting that psychological factors beyond weight status contribute to disordered eating risk in PCOS populations.

TABLE II DISTRIBUTION OF ANNOTATED CONSTRUCTS

| Construct | Train (n=700) | Validation (n=150) | Test (n=150) | Overall (N=1,000) |
|---|---|---|---|---|
| **Body Image Distress** | | | | |
| *Present* | 157 (22.4%) | 31 (20.7%) | 36 (24.0%) | 224 (22.4%) |
| *Absent* | 543 (77.6%) | 119 (79.3%) | 114 (76.0%) | 776 (77.6%) |
| **Disordered Eating** | | | | |
| *Present* | 143 (20.4%) | 31 (20.7%) | 30 (20.0%) | 204 (20.4%) |
| *Absent* | 557 (79.6%) | 119 (79.3%) | 120 (80.0%) | 796 (79.6%) |
| **Metabolic Challenges** | | | | |
| *Present* | 262 (37.4%) | 59 (39.3%) | 55 (36.7%) | 376 (37.6%) |
| *Absent* | 438 (62.6%) | 91 (60.7%) | 95 (63.3%) | 624 (62.4%) |

We operationalized three primary constructs based on validated clinical instruments used in Lee et al.'s study [2]. Body Image Distress was defined using the Eating Disorder Examination-Questionnaire (EDE-Q) shape concern and weight concern subscales, encompassing preoccupation with body shape/weight interfering with daily functioning, importance of shape/weight to self-worth ("number on scale determines my value"), body perception distortion (discrepancies between perceived and actual body size), PCOS-specific concerns (distress about masculinization features, hirsutism, acne, PCOS-related body fat distribution, hair loss), and functional impairment (avoidance behaviors, declining social events, refusing photographs). Lee et al. found strong inverse correlations between eating disorder scores and health-related quality of life across all domains [2]. Disordered Eating Behaviors were defined using DSM-5 criteria, encompassing Binge Eating Disorder (BED), Bulimia Nervosa (BN), restrictive eating disorder (severe caloric restriction under 1000-1200 calories, extended fasting, intense fear of weight gain), Night Eating Syndrome (NES), and orthorexia (obsession with "clean" eating). Women with PCOS showed significantly elevated prevalence of BED (17.6%), BN (6.1%), and NES (12.9%) compared to general population rates, along with more frequent binge episodes and compulsive exercise [2]. Metabolic and Weight Management Challenges were defined recognizing the association between PCOS and obesity, insulin resistance, and type II diabetes mellitus, as well as evidence that individuals with BED experience less weight loss, more rapid weight regain, and higher attrition from weight loss treatments [2]. This construct included weight loss resistance (inability to lose weight despite verified caloric deficit for 2 or more months), weight cycling (3 or more cycles of repeated loss and regain), insulin resistance and metabolic dysfunction, medication-induced changes (birth control causing substantial weight gain, metformin side effects), multiple failed attempts (5 or more different diets/programs without success despite professional involvement), and physical/hormonal barriers (PCOS symptoms, hyperandrogenism, concurrent thyroid issues, metabolic adaptation from extreme dieting history). It is important to note that this construct is operationalized from self-reported text rather than clinical measurements; conditions such as insulin resistance cannot be directly verified from social media posts, and classifications therefore reflect self-reported symptom patterns consistent with metabolic dysfunction rather than confirmed diagnoses.

We developed comprehensive annotation guidelines that provided detailed criteria and differential diagnosis frameworks for each construct. The guidelines emphasized critical distinctions particularly challenging in PCOS populations: body image cognitions versus weight management behaviors (distinguishing thoughts/feelings about body appearance from actions taken for weight loss), disordered eating versus medically supervised dietary management (differentiating pathological restriction from evidence-based PCOS dietary management), and metabolic barriers versus emotional distress about weight (separating genuine metabolic challenges with evidence of resistance from psychological frustration without documented barriers). For each post, annotators provided binary classification (Yes/No) based on explicit criteria thresholds, justification identifying the specific subtype present if Yes or why criteria were not met if No, and key evidence consisting of direct quotes from the post. Guidelines included extensive sections on comorbidity (posts could be marked "Yes" for multiple constructs), PCOS-specific symptom patterns and borderline cases with detailed threshold guidance..

Two annotators (PhD graduate students in Health Informatics) completed a structured training process. After reviewing the Lee et al. [2] study, relevant clinical literature, and comprehensive annotation guidelines (Week 1), annotators independently coded 50 pilot posts representing diverse manifestations of each construct, followed by calibration meetings to discuss discrepancies, clarify guideline interpretations, refine decision rules for PCOS-specific patterns, and establish shared understanding of severity thresholds (Weeks 2-3). Following consensus on pilot cases, annotators independently coded the full 1,000-post dataset without consultation (Weeks 4-8), ensuring that inter-rater reliability metrics would reflect true agreement rather than negotiated consensus. The annotation framework demonstrated strong inter-rater reliability across all three constructs (body image distress: $\kappa$=0.810, disordered eating: $\kappa$=0.852, metabolic challenges: $\kappa$=0.789), supporting the validity of the operationalized criteria and the consistency of their application across annotators.

### C. Fine-tuning Methodology

We employed Low-Rank Adaptation (LoRA) with Rank-Stabilized LoRA (rsLoRA) to enable efficient fine-tuning while preserving pretrained knowledge. LoRA works by injecting trainable low-rank decomposition matrices into the model's attention layers, allowing us to adapt the model with only approximately 1-2% of trainable parameters compared to full fine-tuning [27]. This approach was particularly well-suited to our limited dataset size of 700 training samples, which necessitated careful regularization to prevent overfitting while still capturing the nuanced clinical distinctions required for accurate classification. The rsLoRA variant improves upon standard LoRA by stabilizing training dynamics when using higher ranks [28], which we found necessary for learning the complex, multi-faceted nature of body image distress, disordered eating, and metabolic challenges in PCOS populations. Standard LoRA can exhibit training instability at higher ranks due to scaling issues, but rsLoRA addresses this through adaptive scaling factors that maintain consistent gradient magnitudes throughout training.

During fine-tuning, we optimized cross-entropy loss, the standard objective for training language models to generate text. Unlike traditional classification tasks where models output only class labels, our approach trains models to generate complete structured responses that include both classification decisions (YES/NO for each construct) and detailed justifications with cited evidence. The cross-entropy loss measures how well the model's predicted probability distribution over possible next tokens matches the actual next tokens in the ground truth annotation text at each position in the sequence. Specifically, for each training example, the model receives a post and must generate a multi-paragraph response containing: (1) binary determinations for body image distress, disordered eating, and metabolic challenges, (2) construct-specific subtypes and clinical reasoning, and (3) quoted phrases from the source post as supporting evidence. The loss is calculated across all tokens in this generated response, meaning it encompasses errors in classification decisions (generating "YES" when the ground truth is "NO"), inappropriate clinical terminology (generating "binge eating disorder" when ground truth specifies "restrictive eating"), and inaccurate evidence citation (generating quotes that do not appear in the annotator-provided evidence text). Training loss reflects model performance on the 700 training posts, while validation loss reflects performance on the held-out 150 validation posts that the model never sees during training. The validation loss estimates how well the model will generate accurate, well-justified annotations for truly novel posts. The difference between validation and training loss (the generalization gap) indicates whether the model has memorized training examples or learned generalizable patterns for clinical reasoning. While loss metrics provide continuous feedback during training and enable model selection (choosing the checkpoint with lowest validation loss), they do not directly reveal whether generated classifications are correct or justifications are clinically appropriate. Therefore, we evaluate final model performance primarily through post-hoc analysis: extracting classification decisions from generated text and comparing them to ground truth labels (exact match accuracy, per-label accuracy) and analyzing the quality of generated justifications (citation coverage, span matching, appropriateness assessment).

Our hyperparameter configuration was designed specifically for the challenges of fine-tuning on a small, highly specialized clinical dataset, with each choice reflecting a careful balance between model capacity and generalization. We selected a conservative learning rate of 5e-6, which is significantly lower than typical fine-tuning rates for large language models. This conservative approach follows best practices for instruction-tuning and domain-specific fine-tuning of LLMs, where lower learning rates help prevent overfitting on small datasets and preserve the model's pretrained capabilities [27,29]. This gradual adaptation was particularly important given that clinical posts contain both domain-specific content related to PCOS symptoms and general language patterns, allowing the model to learn clinical nuances while maintaining its broader linguistic capabilities. We trained for an extended duration of 20 epochs, longer than the typical 3-5 epochs used for larger datasets. With only 700 training samples and small batch size, each epoch provides limited gradient updates, so extended training allows the model to fully converge. This was combined with early stopping (patience of 5 evaluation intervals) to halt training if validation loss stopped improving, preventing overfitting despite the many epochs. The extended training also ensures the model sees sufficient examples of rare patterns such as severe body dysmorphia or night eating syndrome, which may appear infrequently in the dataset.

Our choice of LoRA rank ($r = 64$) was at the higher end of the typical range of 8-64 [30], reflecting the complexity of our task. The three constructs involve complex, overlapping clinical distinctions that require higher model capacity, including distinguishing body image cognitions from weight management behaviors, differentiating medically supervised PCOS diets from restrictive eating disorders, and recognizing PCOS-specific patterns such as hirsutism distress and insulin resistance barriers. The higher rank allows learning these nuanced patterns without requiring full fine-tuning [27], and rsLoRA makes this higher rank stable and reliable during training. We applied strong regularization through dropout [30] at a rate of 0.05 and L2 weight decay [31] of 0.01 to prevent memorization of specific training posts and encourage learning of generalizable patterns. Dropout randomly sets 5% of activations to zero during each training step, while weight decay penalizes large parameter values through L2 regularization. This regularization was critical with our dataset of 700 samples where overfitting risk is high, forcing the model to rely on robust features rather than spurious correlations. Additionally, we used gradient accumulation over 4 steps with a batch size of 1, creating an effective batch size of 4. This provides more stable gradient estimates and improves convergence by averaging gradients over multiple samples, while simulating the training dynamics of larger batch sizes without requiring additional GPU memory. Our configuration included the AdamW optimizer [32] with decoupled weight decay and a learning rate schedule with linear warmup for the first 20% of training steps followed by cosine decay. The warmup phase prevents early training instability by gradually increasing the learning rate, while cosine decay provides smooth learning rate reduction to aid convergence.

Models were trained on NVIDIA RTX A6000 GPUs with 50GB VRAM, which provided sufficient memory for batch processing and gradient accumulation while maintaining reasonable training times of approximately 2-3 hours per model. We implemented early stopping based on validation loss with patience of 5 evaluation intervals to prevent overfitting, meaning training would halt if validation loss did not improve for 5 consecutive evaluations, even if training loss continued to decrease. Training dynamics were continuously monitored through several mechanisms. We evaluated validation loss every 50 steps, allowing early detection of overfitting or training instability. Model checkpoints were saved at the same 50-step intervals, preserving intermediate model states and enabling recovery if training degraded or rollback to pre-overfitting checkpoints. The final model for each architecture was selected based on the checkpoint with the lowest validation loss across all training epochs, rather than simply using the last epoch, ensuring we captured the point of optimal generalization.

### D. *Performance Measures*

We evaluated model performance using multiple complementary metrics to assess both individual construct detection and comorbidity identification, alongside systematic explainability analysis. For classification performance, exact match accuracy measured the proportion of posts where the model's predictions perfectly matched ground truth across all three constructs (body image distress, disordered eating, and metabolic challenges), representing a stringent metric that requires complete agreement on all labels simultaneously. To assess differential performance across condition types, we calculated per-label accuracy separately for each of the three constructs, evaluating whether the model correctly classified presence or absence of each specific construct regardless of performance on other constructs. We conducted stratified performance analysis by calculating accuracy separately for posts with varying diagnostic complexity: no conditions present (n=59), single condition present (n=67), and multiple conditions present (n=24), revealing how model performance varies with comorbidity complexity. For comorbidity detection, we calculated Pearson and Spearman correlations between the true number of conditions per post and the model-predicted number of conditions, assessing whether models accurately track the burden of co-occurring conditions independent of which specific constructs are present. Finally, we characterized error patterns by examining false positive rates (no condition incorrectly assigned labels) and false negative rates (with conditions incorrectly predicted as having none), providing insight into whether models err toward over-sensitivity or under-sensitivity.

To assess the transparency and clinical grounding of model predictions, we conducted systematic post-hoc explainability analysis examining how models substantiated their classifications with evidence from the original post text. Our fine-tuning approach required models to generate structured justifications that cite specific phrases from the source post as supporting evidence for each classification decision. We evaluated whether these citations were accurate, comprehensive, and traceable to the original text through an automated pipeline to extract and validate textual evidence. We parsed each model-generated prediction to identify all quoted phrases, which models enclosed in quotation marks as supporting evidence for their classifications, then employed sequence matching to locate each quoted phrase within the original post text. For each successfully matched quote, we recorded the span location (start and end positions), span length in tokens, and whether the match was exact or approximate.

We calculated citation coverage as the proportion of source text tokens cited as evidence in model predictions, operationalized as the percentage of unique tokens from the original post that appeared within any matched span: (number of unique tokens within all matched spans / total tokens in original post) times 100. This metric reveals the breadth of textual evidence models considered when making classifications, with higher coverage percentages indicating models drew upon larger portions of post content. We quantified four key aspects of citation behavior across all test set posts. First, we counted the number of quoted phrases per post in model predictions, reflecting how many distinct pieces of evidence the model attempted to cite regardless of whether they were successfully matched to source text. Second, we counted the number of successfully matched spans in source text, indicating how many quoted phrases were verifiable and traceable to the original post. The ratio of matched spans to quoted phrases serves as a measure of citation accuracy, with higher ratios indicating fewer hallucinated or misattributed quotes. Third, we calculated the average length of matched spans in tokens, revealing whether models cited brief key phrases (2-4 tokens) versus longer contextual passages (10+ tokens). Fourth, we computed the Pearson correlation between the number of quoted phrases and the number of matched spans per post to verify citation accuracy at the post level, with high positive correlations indicating that models consistently cite verifiable evidence rather than generating spurious quotations.

## III. Results

### A. *Inter-Annotator Agreement*

Inter-rater reliability was assessed using Cohen's kappa for binary classifications. The annotation framework demonstrated strong reliability across all constructs: body image distress (κ=0.810), disordered eating (κ=0.852), and metabolic challenges (κ=0.789), with overall agreement for any construct present at κ=0.744. Raw agreement percentages were 93.1%, 95.0%, 90.1%, and 87.9% respectively.

Disagreements comprised 218 instances (6.9%, 5.0%, and 9.9% for body image distress, disordered eating, and metabolic challenges respectively), predominantly involving borderline threshold cases (e.g., weight plateaus lasting 6-8 weeks vs. the required 3+ months, caloric intake of 1200-1400 calories vs. the under 1000 calorie restriction threshold) and PCOS-specific boundary situations (e.g., distinguishing medical dietary management from pathological restriction). Annotator 2 applied slightly more inclusive criteria (body image distress: 25.8% vs. 21.9%; disordered eating: 22.8% vs. 20.2%; metabolic challenges: 38.3% vs. 36.6%), resulting in 150 false positives vs. 68 false negatives.

For fine-tuning, we adopted an inclusive labeling strategy where either annotator marking a construct as present resulted in positive classification. This approach prioritized sensitivity to capture the full spectrum of construct manifestations, particularly borderline presentations, ensuring comprehensive representation of psychological experiences in PCOS populations. Given the high inter-annotator agreement (κ=0.789-0.852), disagreements constituted only 5.0-9.9% of posts per construct.

### B. *Model Performance Comparison*

We evaluated three small language models: DeepSeek-R1-Distill-Qwen-1.5B, Qwen3-1.7B, and Gemma-2-2B. All models were fine-tuned using identical hyperparameters and the same 700-post training dataset to generate structured annotations following our guidelines.

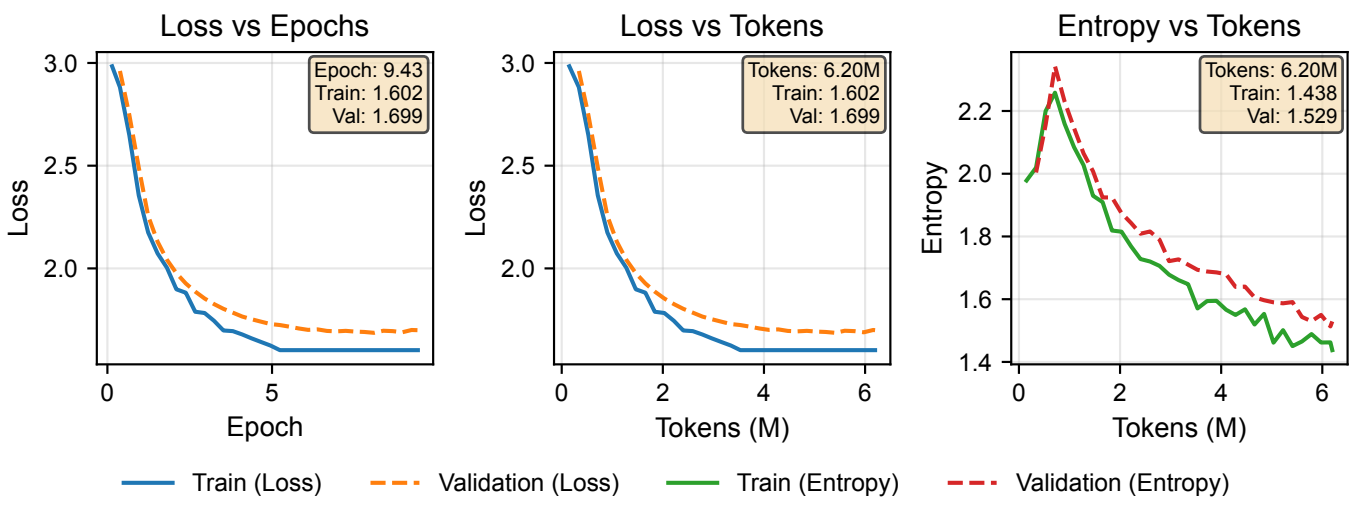

**Fig. 1.** Training and validation loss curves for DeepSeek-R1-Distill-Qwen-1.5B.

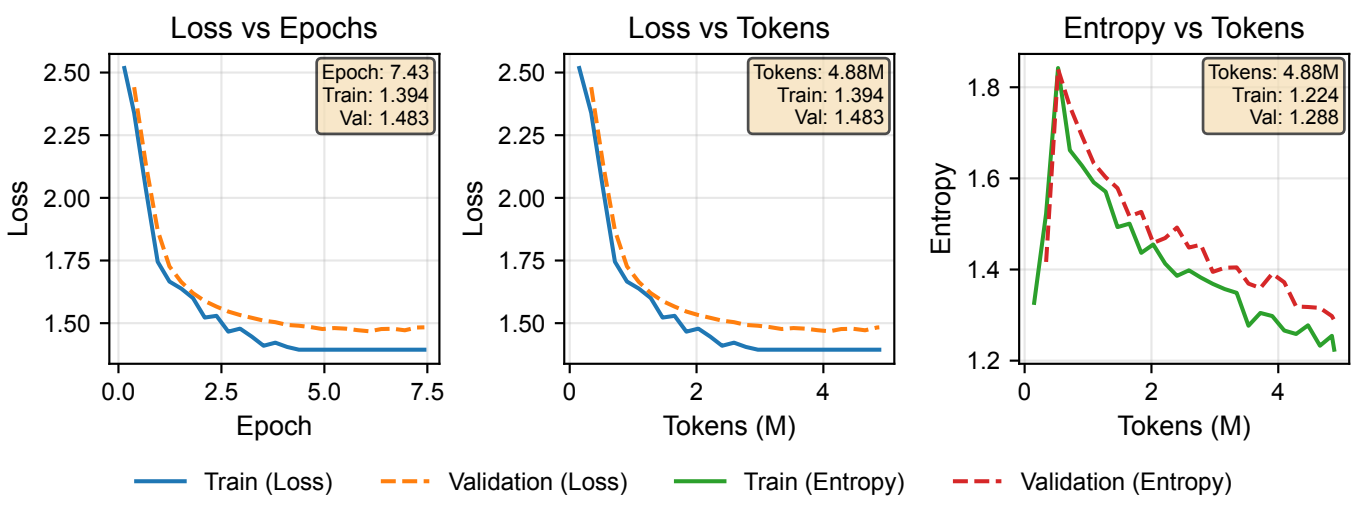

**Fig. 2.** Training and validation loss curves for Qwen3-1.7B.

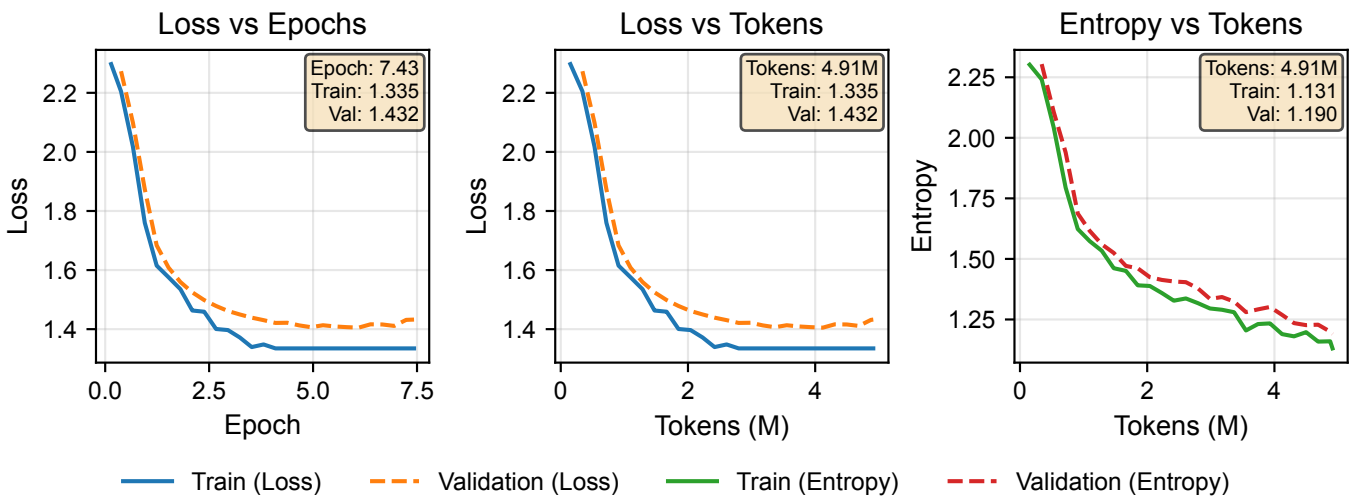

**Fig. 3**. Training and validation loss curves for Gemma-2-2B.

The models exhibited distinct convergence patterns during fine-tuning (Figs. 1-3). DeepSeek-R1-Distill-Qwen-1.5B required the most training (6.20M tokens, 9.43 epochs), suggesting difficulty learning nuanced clinical distinctions. Qwen3-1.7B and Gemma-2-2B converged 25% faster (4.9M tokens, 7.43 epochs), indicating more efficient adaptation.

Gemma-2-2B achieved the best performance with lowest training loss (1.3349) and validation loss (1.4324), most accurately replicating expert annotations. Qwen3-1.7B performed competitively (training: 1.3942, validation: 1.4834), while DeepSeek-R1-Distill-Qwen-1.5B struggled with higher losses (training: 1.6021, validation: 1.6993). All models showed modest generalization gaps (0.0892-0.0976), indicating effective regularization without overfitting.

Entropy metrics revealed annotation confidence patterns. Gemma-2-2B demonstrated highest confidence with lowest entropy (training: 1.1305, validation: 1.1900) and developed clear decision boundaries. Qwen3-1.7B showed moderate confidence (training: 1.2240, validation: 1.2885), while DeepSeek-R1-Distill-Qwen-1.5B exhibited greater uncertainty (training: 1.4376, validation: 1.5291). Small entropy gaps across validation and training sets (0.0595-0.0915) indicated consistent annotation criteria across datasets.

## C. *Detection of Multiple Conditions*

The models accurately identified the prevalence of comorbidity in the dataset. This close alignment (Table III) for multiple conditions (16.0-17.3%) suggests the models are well-calibrated for detecting comorbidity frequency, neither systematically over-identifying nor under-identifying co-occurring conditions.

TABLE III COMPARISON OF CO-OCCURRING CONDITION DISTRIBUTIONS

| **Distribution** | **Ground Truth** | **Gemma-2-2B** | **DeepSeek-R1-Distill-Qwen-1.5B** | **Qwen3-1.7B** |
|---|---|---|---|---|
| *0 conditions* | 39.3% | 33.3% | 32.0% | 27.3% |
| *1 condition* | 44.7% | 49.3% | 52.0% | 56.0% |
| *2+ conditions* | 16.0% | 17.3% | 16.0% | 16.7% |

TABLE IV MODEL ACCURACY BY NUMBER OF CO-OCCURRING CONDITIONS

| **Diagnosis Level (n)** | **Gemma-2-2B** | **DeepSeek-R1-Distill-Qwen-1.5B** | **Qwen3-1.7B** |
|---|---|---|---|
| **No diagnosis (59)** | 78.0% | 71.2% | 66.1% |
| **Single diagnosis (67)** | 74.6% | 80.6% | 79.1% |
| **Multiple diagnoses (24)** | 70.6% | 66.7% | 66.7% |
| **Overall (150)** | 75.3% | 74.7% | 72.0% |

Table IV summarizes exact match accuracy across diagnostic complexity levels. Examining individual constructs revealed differential sensitivity to comorbidity.

TABLE V CONSTRUCT-LEVEL DETECTION ACCURACY

| **Construct** | **Context** | **Gemma-2-2B** | **DeepSeek-R1-Distill-Qwen-1.5B** | **Qwen3-1.7B** |
|---|---|---|---|---|
| **Body Image** | *Single* | 11/13 (84.6%) | 11/13 (84.6%) | 11/13 (84.6%) |
| | *Multiple* | 15/20 (75.0%) | 16/20 (80.0%) | 17/20 (85.0%) |
| **Disordered Eating** | | | | |
| | *Single* | 11/13 (84.6%) | 12/13 (92.3%) | 12/13 (92.3%) |
| | *Multiple* | 11/12 (91.7%) | 10/12 (83.3%) | 11/12 (91.7%) |
| **Metabolic Challenges** | | | | |
| | *Single* | 35/41 (85.4%) | 36/41 (87.8%) | 35/41 (85.4%) |
| | *Multiple* | 15/20 (75.0%) | 14/20 (70.0%) | 14/20 (70.0%) |

Table V presents per-label accuracy stratified by diagnostic complexity, showing both correct predictions and total cases.

TABLE VI PRECISION, RECALL AND F1 SCORES ACROSS MODELS

| **Label** | **Metric** | **Gemma-2-2B** | **DeepSeek-R1-Distill-Qwen-1.5B** | **Qwen3-1.7B** |
|---|---|---|---|---|
| **Body Image** | *Precision* | 0.76 | 0.71 | 0.69 |
| | *Recall* | 0.74 | 0.73 | 0.78 |
| | *F1* | 0.75 | 0.72 | 0.73 |
| **Disordered Eating** | | | | |
| | *Precision* | 0.77 | 0.73 | 0.75 |
| | *Recall* | 0.78 | 0.75 | 0.78 |
| | *F1* | 0.77 | 0.74 | 0.76 |
| **Metabolic Challenges** | | | | |
| | *Precision* | 0.73 | 0.70 | 0.68 |
| | *Recall* | 0.76 | 0.74 | 0.74 |
| | *F1* | 0.74 | 0.72 | 0.71 |

Table VI presents per-label precision, recall, and F1 scores across all three models. Gemma-2-2B achieved the strongest overall performance, with F1 scores of 0.75, 0.77, and 0.74 for body image distress, disordered eating, and metabolic challenges respectively. DeepSeek-R1-Distill-Qwen-1.5B and Qwen3-1.7B performed comparably across most constructs, with F1 scores ranging from 0.72-0.74 and 0.71-0.76 respectively. Disordered eating achieved the highest or jointly highest F1 scores across all models (0.74-0.77), while metabolic challenges, despite its higher prevalence in the dataset, showed more variable F1 performance (0.71-0.74), likely reflecting greater difficulty in classifying this construct in comorbid contexts. Body image distress showed the lowest F1 scores for Gemma-2-2B and DeepSeek-R1-Distill-Qwen-1.5B (0.72-0.75), consistent with its more subjective and nuanced presentation. Notably, Qwen3-1.7B achieved the highest recall for body image distress (0.78) despite the lowest precision (0.69) across all three models, suggesting a tendency toward over-identification of this construct, which aligns with its higher false positive rate observed in the comorbidity analysis.

Table VII presents the correlation between true and predicted label counts, with strong agreement across all models.

TABLE VII CORRELATION BETWEEN PREDICTED AND ACTUAL NUMBER

| Metric | Gemma-2-2B | DeepSeek-R1-Distill-Qwen-1.5B | Qwen3-1.7B |
|---|---|---|---|
| **Pearson r** | 0.782*** | 0.730*** | 0.733*** |
| **Spearman ρ** | 0.777*** | 0.721*** | 0.737*** |

*Note: *** p < 0.001*

All three models demonstrated strong capability in detecting co-occurring conditions, with Pearson correlations ranging from 0.730-0.782 (all $p < 0.0001$), indicating that predicted label counts closely tracked true label counts. Gemma showed the strongest correlation ($r = 0.782$), while DeepSeek-R1-Distill-Qwen-1.5B and Qwen3-1.7B performed comparably ($r = 0.730$ and 0.733, respectively).

Analysis of error patterns (Figure 4) revealed systematic biases in model predictions that favor sensitivity over specificity. False positives (under-prediction) were most common for posts with no diagnosis, with models incorrectly assigning one or more labels to 22.0-33.9% of posts that contained no target constructs. DeepSeek-R1-Distill-Qwen-1.5B incorrectly assigned labels to 17 of 59 posts with no condition (28.8%), Gemma-2-2B to 13 of 59 posts (22.0%), and Qwen3-1.7B to 20 of 59 posts (33.9%). False negatives were minimal across all models, with DeepSeek-R1-Distill-Qwen-1.5B and Gemma-2-2B each incorrectly predicting only 1 post with 2 true conditions as having none, while Qwen3-1.7B produced 0 such errors. This asymmetry suggests the models err on the side of sensitivity rather than specificity, which may be clinically appropriate for screening applications where false negatives carry greater risk.

For comorbidity capture in posts with 2 or more conditions (n=24), models demonstrated varying capability in recognizing complex, multi-faceted clinical presentations. DeepSeek-R1-Distill-Qwen-1.5B correctly identified 15 of 24 posts (62.5%) as having 2 or more conditions, Gemma-2-2B correctly identified 18 of 24 posts (75.0%), and Qwen3-1.7B correctly identified 16 of 24 posts (66.7%). Gemma-2-2B's superior performance of 75% indicates better capability for recognizing complex presentations where multiple constructs co-occur.

## D. Explainability Analysis

All three fine-tuned models demonstrated substantial evidence-based reasoning, with every post (150/150, 100%) containing at least one matched span between the model's prediction and the original text (Table VIII). The models averaged between 11.4-11.8 matched text spans per post, derived from 13.2-15.0 quoted phrases in their predictions. The average matched span length ranged from 6.3-7.8 tokens.

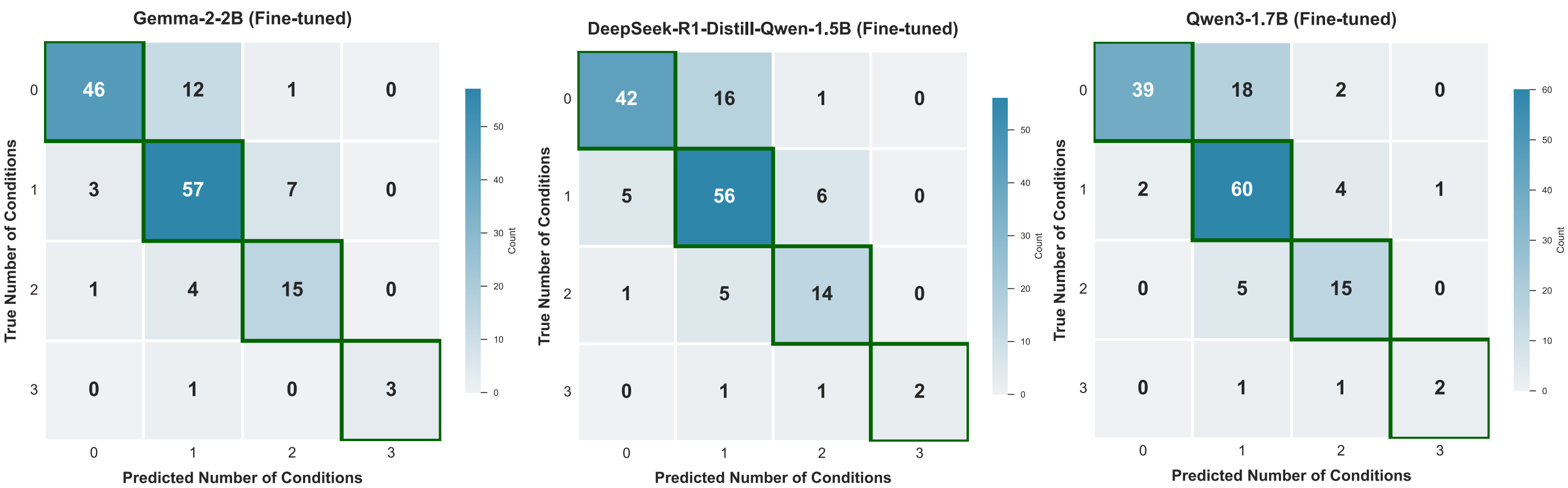

**Fig. 4.** Predicted vs. actual co-occurrence levels (a) Gemma-2-2B (b) DeepSeek-R1-Distill-Qwen-1.5B (c) Qwen3-1.7B as confusion matrices

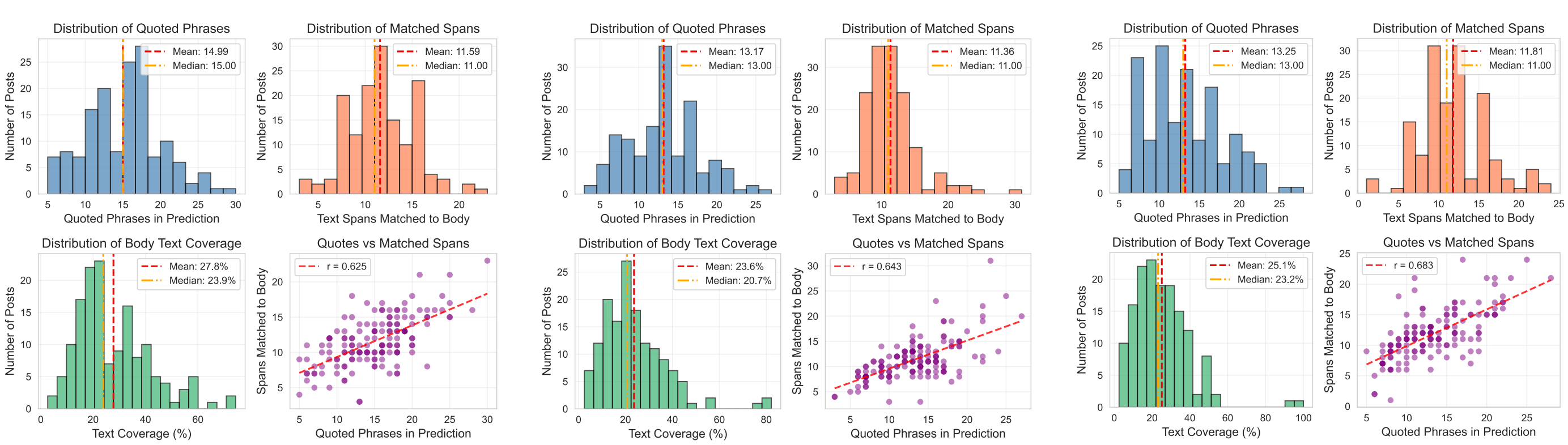

**Fig. 5.** Source text coverage by evidence spans for each model. (a) Gemma-2-2B. (b) DeepSeek-R1-Distill-Qwen-1.5B. (c) Qwen3-1.7B

TABLE VIII EVIDENCE GROUNDING VIA TEXT SPAN MATCHING

| **Metric** | **Gemma-2-2B** | **DeepSeek-R1-Distill-Qwen-1.5B** | **Qwen3-1.7B** |
|---|---|---|---|
| **Total Posts** | 150 | 150 | 150 |
| **Posts & Matched Spans, n(%)** | 150 (100%) | 150 (100%) | 150 (100%) |
| **Avg. Quoted Phrases per Post** | 14.99 | 13.17 | 13.25 |
| **Avg. Spans Matched to Body per Post** | 11.59 | 11.36 | 11.81 |
| **Avg. Span Length (words)** | 7.8 | 6.3 | 6.4 |
| **Avg. Body Text Coverage (%)** | 27.8% | 23.6% | 25.1% |

Coverage analysis revealed that models grounded their predictions in 23.6%-27.8% of the source text tokens (Figure 5). Gemma-2-2B showed the highest text coverage (27.8%), followed by Qwen3-1.7B (25.1%) and DeepSeek-R1-Distill-Qwen-1.5B (23.6%). The correlation between quoted phrases and matched spans (r > 0.62 for all models demonstrates that model citations were accurately traceable to the source material, supporting the reliability of model-generated justifications.

### *E. Illustrative Example of Evidence Grounding*

To demonstrate how models ground their predictions in source text, we present a representative example (Post Index 60) where a user described frustration with medication-related weight gain despite healthy behaviors. Figures 6 to 8 show the span analysis for all three models. The original post stated (first 100 words): "*I'm super frustrated about my weight gain... between 2020-2023 I lost almost 15 kgs... However on September 2023 I was diagnosed with PCOS and started my antidepressants. This when I started gaining weight uncontrollably although I was very active and I was eating well. Two months ago I started going to the gym more seriously and made sure to eat protein...*"

All models correctly classified this as metabolic and weight management challenges but demonstrated different evidence extraction patterns. Gemma-2-2B identified 8 spans covering 34.3% of the text (18 quoted phrases, average span, 7.8 tokens), such as "I'm super frustrated about my weight gain”.

**Original Post (first 100 words):**

Hey guys , I'm super frustrated about my weight gain . Here's a summary of how it started to happen : Basically between 2020-2023 I lost almost 15, kgs , I wouldn't say in the healthiest way but I was in a cycle of mostly restricting and bingeing from time to time . However on September 2023 I was diagnosed with pcos and started my antidepressants. This when I started gaining weight uncontrollably although I was very active and I was eating well. Two months ago I started going to the gym more seriously and made sure to eat protein...

```
Span Analysis Summary:
Quoted phrases: 18  |  Matched spans: 8
Coverage: 34.3%  |  Avg span length: 7.8 tokens
```

**Example Evidence Spans (showing 8 of 8):**

```
Span 1: Q: "I'm super frustrated about my weight gain"
        M: "i'm super frustrated about my weight gain"  [pos 2-9, len=7]
Span 2: Q: "I don't know what I'm doing wrong but I need advices on how ..."
        M: "i don't know what i'm doing wrong but i need advices on how ..."  [pos 118-136, len=18]
Span 3: Q: "eating well"
        M: "eating well"  [pos 77-79, len=2]
Span 4: Q: "eating protein rich food,"
        M: "protein rich food"  [pos 95-98, len=3]
Span 5: Q: "made sure to eat protein rich food,"
        M: "made sure to eat protein rich food"  [pos 91-98, len=7]
Span 6: Q: "gym more seriously"
        M: "gym more seriously"  [pos 87-90, len=3]
Span 7: Q: "This when I started gaining weight uncontrollably"
        M: "this when i started gaining weight uncontrollably"  [pos 62-69, len=7]
Span 8: Q: "Started going to the gym more seriously and made sure to eat..."
        M: "started going to the gym more seriously and made sure to eat..."  [pos 83-98, len=15]
```

**Fig. 6.** Matched text spans by Gemma-2-2B in Post 60

**Original Post (first 100 words):**

Hey guys , I'm super frustrated about my weight gain . Here's a summary of how it started to happen : Basically between 2020-2023 I lost almost 15, kgs , I wouldn't say in the healthiest way but I was in a cycle of mostly restricting and bingeing from time to time . However on September 2023 I was diagnosed with pcos and started my antidepressants. This when I started gaining weight uncontrollably although I was very active and I was eating well. Two months ago I started going to the gym more seriously and made sure to eat protein...

```
Span Analysis Summary:
Quoted phrases: 17  |  Matched spans: 10
Coverage: 32.9%  |  Avg span length: 8.2 tokens
```

**Example Evidence Spans (showing 10 of 10):**

```
Span 1: Q: "weight gain,"
        M: "weight gain"  [pos 7-9, len=2]
Span 2: Q: "Gained 4 kgs"
        M: "gained 4 kgs"  [pos 106-109, len=3]
Span 3: Q: "gaining weight despite efforts"
        M: "gaining weight"  [pos 66-68, len=2]
Span 4: Q: "eating well"
        M: "eating well"  [pos 77-79, len=2]
Span 5: Q: "gained 4 kgs [uncontrollably]"
        M: "gained 4 kgs"  [pos 106-109, len=3]
Span 6: Q: "started going to the gym more seriously and made sure to eat..."
        M: "started going to the gym more seriously and made sure to eat..."  [pos 83-109, len=26]
Span 7: Q: "Lost almost 15 kgs... then gained 4 kgs [uncontrollably]"
        M: "lost almost 15 kgs"  [pos 23-27, len=4]
Span 8: Q: "Lost almost 15 kgs... then gained 4 kgs [uncontrollably]"
        M: "gained 4 kgs"  [pos 106-109, len=3]
Span 9: Q: "Diagnosed with PCOS and started my antidepressants... two mo..."
        M: "diagnosed with pcos and started my antidepressants"  [pos 55-62, len=7]
Span 10: Q: "Diagnosed with PCOS and started my antidepressants... two mo..."
        M: "two months ago i started going to the gym more seriously and..."  [pos 79-109, len=30]
```

**Fig. 7.** Matched text spans by DeepSeek-R1-Distill-Qwen-1.5B in Post 60

**Original Post (first 100 words):**

Hey guys , I'm super frustrated about my weight gain . Here's a summary of how it started to happen : Basically between 2020-2023 I lost almost 15, kgs , I wouldn't say in the healthiest way but I was in a cycle of mostly restricting and bingeing from time to time . However on September 2023 I was diagnosed with pcos and started my antidepressants. This when I started gaining weight uncontrollably although I was very active and I was eating well. Two months ago I started going to the gym more seriously and made sure to eat protein...

```
Span Analysis Summary:
Quoted phrases: 21  |  Matched spans: 15
Coverage: 38.5%  |  Avg span length: 4.9 tokens
```

**Example Evidence Spans (showing 10 of 15):**

```
Span 1: Q: "frustrated about my weight gain,"
        M: "frustrated about my weight gain"  [pos 4-9, len=5]
Span 2: Q: "need advice on how to drop this weight"
        M: "on how"  [pos 129-131, len=2]
Span 3: Q: "need advice on how to drop this weight"
        M: "drop this weight"  [pos 133-136, len=3]
Span 4: Q: "Frustrated about my weight gain"
        M: "frustrated about my weight gain"  [pos 4-9, len=5]
Span 5: Q: "Need advices on how I can drop this weight"
        M: "need advices on how i can drop this weight"  [pos 127-136, len=9]
Span 6: Q: "gaining weight,"
        M: "gaining weight"  [pos 66-68, len=2]
Span 7: Q: "eating protein-rich foods"
        M: "protein rich"  [pos 95-97, len=2]
Span 8: Q: "diagnosed with PCOS,"
        M: "diagnosed with pcos"  [pos 55-58, len=3]
Span 9: Q: "restricting and bingeing from time to time"
        M: "restricting and bingeing from time to time"  [pos 42-49, len=7]
Span 10: Q: "lost almost 15, kgs... mostly restricting and bingeing from ..."
        M: "lost almost 15 kgs"  [pos 23-27, len=4]
```

**Model: Qwen3-1.7B (Fine-tuned) | Post Index: 60**

**Fig. 8.** Matched text spans by Qwen3-1.7B in Post 60

DeepSeek-R1-Distill-Qwen-1.5B extracted 10 distinct spans covering 32.9% of the text (17 quoted phrases, average span length 8.2 tokens), citing key phrases such as "weight gain," "diagnosed with PCOS and started my antidepressants," "gaining weight uncontrollably," and "started going to the gym more seriously and made sure to eat protein rich food." Notably, DeepSeek-R1-Distill-Qwen-1.5B identified one longer contextual span of 30 tokens capturing the complete narrative of dietary efforts and continued weight gain.

Qwen3-1.7B extracted the most spans (15 spans covering 38.5% of text, 21 quoted phrases) but with shorter average length (4.9 tokens), including "frustrated about my weight gain," "diagnosed with PCOS," and "restricting and bingeing from time to time."

## IV. Discussion

This study demonstrates that small, open-source language models (under 2B parameters) can reliably detect body image distress, disordered eating behaviors, and metabolic challenges in PCOS-related social media posts with clinically grounded explainability. Our fine-tuned models achieved exact match accuracies of 72.0-75.3% on held-out test data, with robust detection of co-occurring conditions (Pearson $r = 0.730$-$0.782$, $p < 0.0001$) and accurate comorbidity prevalence estimation (16.0-17.3% vs. ground truth 16.0%). Critically, all predictions included traceable evidence from source text, with models grounding classifications in 23.6-27.8% of post content through 11.4-11.8 cited text spans per post. These findings address three critical gaps in mental health NLP: reliance on large proprietary models lacking transparency, focus on single-condition detection rather than comorbid symptomatology, and absence of clinically grounded explainability aligned with established diagnostic frameworks. Differential performance across constructs in comorbid contexts reveals that disordered eating detection remained robust or improved in comorbid presentations (Gemma-2-2B: 84.6% single vs. 91.7% multiple), while metabolic challenges showed the largest decline (70.0-75.0% vs. 85.4-87.8% in single-condition cases).

The explainability analysis revealed 100% of predictions contained traceable citations, with strong correlation between quoted phrases and matched spans ($r > 0.62$). This transparency builds clinical trust as reasoning is expressed in clinically meaningful terms rather than statistical associations.

Accurate detection of co-occurring conditions has direct implications for treatment planning. As Lee et al. documented, eating disorder symptoms correlate inversely with health-related quality of life ($r = -0.57$, $p < 0.001$), and individuals with binge eating disorder experience less weight loss, more rapid weight regain, and higher treatment attrition [2]. These findings are consistent with a recent large-scale systematic review and meta-analysis of 20 studies involving over 28,000 women with PCOS, which found 1.5 to 2.9 times higher odds of BED, BN, and disordered eating in women with PCOS compared to controls, with elevated risk persisting regardless of BMI status [1]. Also, current PCOS guidelines emphasize weight loss for metabolic management, which can be risky with ED [4]. Our models' capability to simultaneously detect body image distress, eating disorders, and metabolic challenges could inform more nuanced treatment planning, enabling clinicians to prioritize interventions and better tailor treatment intensity.

Several limitations warrant consideration. First, our dataset derived from Reddit communities focused on eating disorders and weight management may not represent the full PCOS spectrum, with 60.5% of posts exhibiting at least one target construct. Second, requiring explicit "PCOS" mention excludes individuals who discuss symptoms without naming the condition. Third, our annotation operationalized clinical constructs from self-reported symptoms rather than clinical interviews, meaning models detect self-reported symptom patterns rather than confirmed diagnoses. Fourth, metabolic challenges performance in comorbid contexts (70.0-75.0%) indicates room for improvement. Fifth, we did not conduct formal validation of citation quality by clinical experts. Finally, we did not evaluate performance across demographic subgroups, though PCOS presentations may vary.

This work demonstrates that small, open-source language models can reliably detect the PCOS triple burden with clinically grounded explainability, achieving screening-suitable performance (72.0-75.3% exact match accuracy) while maintaining full transparency and deployability in resource-constrained settings. The performance gap between single and multiple diagnoses (4.0-13.9 percentage points) indicates these models should serve as decision support tools rather than autonomous diagnostic systems, with human clinical expertise remaining essential for complex comorbid presentations. By demonstrating that compact, transparent models can tackle the complex task of detecting the PCOS triple burden with clinically meaningful explainability, this work provides a foundation for more accessible, ethical, and effective use of artificial intelligence in mental health screening and support for chronically ill populations. Future work should focus on three directions: prospective validation in real clinical settings such as PCOS specialty clinics to establish practical utility; expansion to more demographically and linguistically diverse datasets and platforms beyond Reddit to improve generalizability; and integration with patient-facing applications such as symptom trackers or clinical chatbots to enable continuous, low-burden screening that supports earlier identification of at-risk individuals.